\documentclass[format=sigconf]{acmart}

\usepackage{bm}
\usepackage{amsfonts}
\usepackage{amsmath}
\usepackage{color}
\usepackage{graphicx}
\usepackage{subfigure}
\usepackage{booktabs}
\usepackage{multirow}
\usepackage{enumitem}
\usepackage{setspace}
\usepackage{colortbl}
\usepackage{pifont}
\usepackage{balance}
\usepackage{xspace}
\usepackage{array}
\usepackage{stfloats}
\usepackage{algpseudocode}
\usepackage[noend,ruled,linesnumbered]{algorithm2e}
\usepackage{makecell}

\definecolor{myblue}{rgb}{0.2, 0.2, 0.9}

\newcommand{\bmh}{\bm h}
\newcommand{\bmu}{\bm u}
\newcommand{\bmv}{\bm v}
\newcommand{\model}{SeedTopicMine\xspace}

\begin{document}
\title{Effective Seed-Guided Topic Discovery by \\ Integrating Multiple Types of Contexts} 

\author{Yu Zhang$^*$}
\author{Yunyi Zhang$^*$}
\affiliation{
\institution{University of Illinois at Urbana-Champaign} 
\country{}
\institution{\{yuz9, yzhan238\}@illinois.edu}
}

\author{Martin Michalski$^*$}
\author{Yucheng Jiang$^*$}
\affiliation{
\institution{University of Illinois at Urbana-Champaign} 
\country{}
\institution{\{martinm6, yj17\}@illinois.edu}
}

\author{Yu Meng$^*$}
\author{Jiawei Han}
\affiliation{
\institution{University of Illinois at Urbana-Champaign} 
\country{}
\institution{\{yumeng5, hanj\}@illinois.edu}
}

\thanks{$^*$Equal Contribution.}

\renewcommand{\shortauthors}{Yu Zhang, Yunyi Zhang, Martin Michalski, Yucheng Jiang, Yu Meng, \& Jiawei Han}

\begin{abstract}
Instead of mining coherent topics from a given text corpus in a completely unsupervised manner, seed-guided topic discovery methods leverage user-provided seed words to extract distinctive and coherent topics so that the mined topics can better cater to the user's interest.
To model the semantic correlation between words and seeds for discovering topic-indicative terms, existing seed-guided approaches utilize different types of context signals, such as document-level word co-occurrences, sliding window-based local contexts, and generic linguistic knowledge brought by pre-trained language models. 
In this work, we analyze and show empirically that each type of context information has its value and limitation in modeling word semantics under seed guidance, but combining three types of contexts (i.e., word embeddings learned from local contexts, pre-trained language model representations obtained from general-domain training, and topic-indicative sentences retrieved based on seed information) allows them to complement each other for discovering quality topics.
We propose an iterative framework, \textsc{\model}, which jointly learns from the three types of contexts and gradually fuses their context signals via an ensemble ranking process.
Under various sets of seeds and on multiple datasets, \textsc{\model} consistently yields more coherent and accurate topics than existing seed-guided topic discovery approaches.
\end{abstract}



\copyrightyear{2023}
\acmYear{2023}
\setcopyright{acmcopyright}\acmConference[WSDM '23]{Proceedings of the Sixteenth ACM International Conference on Web Search and Data Mining}{February 27-March 3, 2023}{Singapore, Singapore}
\acmBooktitle{Proceedings of the Sixteenth ACM International Conference on Web Search and Data Mining (WSDM '23), February 27-March 3, 2023, Singapore, Singapore}
\acmPrice{15.00}
\acmDOI{10.1145/3539597.3570475}
\acmISBN{978-1-4503-9407-9/23/02}

\begin{spacing}{0.965}
\maketitle

\section{Introduction}
To efficiently grasp the information in a large collection of documents, it is of great interest to automatically discover a set of coherent topics from the corpus. Besides capturing meaningful structures in massive text data \cite{griffiths2004finding}, topic discovery also widely benefits downstream text mining tasks such as taxonomy construction \cite{lee2022taxocom} and document classification \cite{chen2015dataless}.

Unsupervised topic models, from LDA \cite{blei2003latent} to embedding-based \cite{dieng2020topic,xun2017collaboratively} and pre-trained language model-enhanced \cite{sia2020tired,meng2022topic} approaches, have been extensively studied for decades as the mainstream approach to topic discovery. Despite their efficacy in uncovering prominent themes of a corpus, such models tend to retrieve semantically general topics that may not align well with users' specific interests, as explained in \cite{meng2020discriminative,harandizadeh2022keyword}. Motivated by this, rather than finding arbitrary topics in a fully unsupervised manner, seed-guided topic discovery \cite{jagarlamudi2012incorporating,gallagher2017anchored,meng2020discriminative,harandizadeh2022keyword,zhang2022seed} aims to extract topics along a certain dimension based on user-provided seeds, and the top-ranked words under each topic should be discriminatively relevant to the corresponding seed. For example, given a collection of restaurant reviews, if a user would like to explore topics of \textit{food types} (e.g., by providing the seeds ``\textit{noodles}'', ``\textit{steak}'', and ``\textit{pizza}''), then a seed-guided topic discovery model should discover topic-indicative terms for each input seed (e.g., ``\textit{ramen}'' and ``\textit{pasta}'' under ``\textit{noodles}''), instead of finding terms that are relevant to multiple seeds (e.g., ``\textit{beef}'') or retrieving topics along other dimensions (e.g., ``\textit{good}'' or ``\textit{bad}'' as topics of \textit{sentiments}).

Recent studies on seed-guided topic discovery \cite{meng2020discriminative,harandizadeh2022keyword,zhang2022seed,lee2022taxocom} have been focusing on utilizing different types of context information so that they can go beyond the ``bag-of-words'' generative assumption in LDA and learn more accurate word semantics for topic discovery. To be specific, there are three major types of context signals used in related studies.

\vspace{1mm}

\noindent \textbf{Skip-Gram Word Embeddings.} Different from LDA which infers topics based on the global document-word frequency matrix, skip-gram embedding learning \cite{mikolov2013distributed} assumes that words occurring in similar local contexts (e.g., $\pm 5$ words) tend to have similar semantic properties. Following this assumption, each word in the corpus can be represented by an embedding vector in a latent space. To incorporate skip-gram signals in topic discovery, the word embeddings can be injected into the LDA backbone \cite{harandizadeh2022keyword} or jointly learned by viewing documents and seeds also as contexts \cite{meng2020discriminative}. However, skip-gram embeddings are less helpful in disambiguating word meanings because only one vector is learned for each word given the whole corpus. Indeed, Sia et al. \cite{sia2020tired} show that clustering skip-gram embeddings underperforms clustering output representations of contextualized language models such as BERT \cite{devlin2019bert} in unsupervised topic modeling.

\vspace{1mm}

\noindent \textbf{Pre-trained Language Model Representations.} Pre-trained language models (PLMs) \cite{devlin2019bert,radford2019language,liu2019roberta} have revolutionized the text mining field by learning contextualized word embeddings. The Transformer architecture \cite{vaswani2017attention} used in many PLMs can capture long-range and high-order context signals, and the knowledge learned by PLMs from web-scale corpora can complement contexts in the input corpus in topic discovery \cite{zhang2022seed}. Meanwhile, related studies have observed several cases where PLMs generate noticeably bad topics. For example, Meng et al. \cite{meng2022topic} show that PLM representations suffer from the curse of dimensionality and do not form clearly separated clusters; Thompson and Mimno \cite{thompson2020topic} find that GPT-2 representations \cite{radford2019language} work well only if the outputs of certain layers are taken, and RoBERTa-induced topics \cite{liu2019roberta} are consistently of poor quality.

\vspace{1mm}

\noindent \textbf{Topic-Indicative Documents.} Although skip-gram embeddings and PLMs are powerful in representing each word based on its contexts, neither of them considers whether the contexts they use are topic-indicative (i.e., semantically close to a certain seed). In fact, skip-gram embedding learning always takes the $\pm x$ words as contexts, regardless of whether they are relevant to any seed; a PLM will always output the same representation for a word if the input corpus is fixed, no matter what the seeds are. To tackle this problem, supervised topic models \cite{mcauliffe2007supervised,lacoste2008disclda} propose to leverage document-level training data (i.e., each document belongs to which seed or semantic category). However, such information relies on massive human annotation, which may be difficult to obtain in practical applications (e.g., weakly supervised text classification \cite{meng2018weakly,mekala2020contextualized,zhang2022motifclass}). Moreover, a document may be too broad to be viewed as a context unit because each document can be relevant to multiple topics simultaneously.

To summarize, each type of context signals has its specific advantages and disadvantages. Therefore, a topic discovery method purely relying on one type of context information may not be robust across different datasets or seed dimensions. Meanwhile, it is worth noting that the three types of contexts strongly complement each other. For example, PLMs have contextualization power which skip-gram embeddings are short of; skip-gram embeddings usually have fewer dimensions than PLM representations and are less prone to the curse of dimensionality; topic-indicative documents are not naturally available, but they can be retrieved by applying skip-gram embeddings and PLMs.

\vspace{1mm}

\noindent \textbf{Contributions.} Motivated by the complementarity of context signals, in this paper, we propose \textsc{\model}, an effective seed-guided topic discovery framework by integrating multiple types of contexts. \textsc{\model} iteratively retrieves and updates the set of topic-discriminative terms for each seed. In each iteration, we first jointly leverage seed-guided skip-gram embeddings and PLM-based representations to discover a set of topic-indicative terms. Then, using these terms, we retrieve a set of topic-indicative sentences. Here, we consider sentences rather than documents because each sentence, as a more fine-grained unit, is more likely to concentrate on one topic. Finally, the derived topic-indicative sentences and the other two types of contexts are cooperatively utilized through an ensemble ranking process, after which the topic-discriminative terms will be updated and used for the next iteration.

Extensive experiments on real-world datasets show that \textsc{\model} effectively discovers discriminative terms under each seed to form coherent topics. Our human evaluation quantitatively validates the superiority of \textsc{\model} over baselines that rely on a single type of contexts. In the ablation study, we observe that even in the same dataset, if we consider different dimensions of seeds, the contributions of different context signals vary significantly, which confirms our key motivation that any single type of context signal is insufficient for discovering seed-discriminative topics stably.

\section{Problem Definition}
Following \cite{meng2020discriminative}, we assume a seed can be either a unigram or a phrase. Given an input corpus and a set of seeds, our goal is to find a set of terms under each seed to form a coherent topic. Conforming to the assumption of seeds, each term can also be a unigram or a phrase. In practice, given a raw corpus, one can adopt existing phrase chunking tools \cite{manning2014stanford,shang2018automated} to obtain phrases in it.

\begin{definition}{(Problem Definition)} 
Given a corpus $\mathcal{D}=\{d_1,...,$ $d_{|\mathcal{D}|}\}$ and a set of seeds $\mathcal{S}=\{s_1,...,s_{|\mathcal{\mathcal{S}}|}\}$, seed-guided topic discovery aims to find a set of terms $\mathcal{T}_i=\{t_{i1},t_{i2},...,t_{i|\mathcal{T}_i|}\}$ appearing in $\mathcal{D}$ for each seed $s_i$ $(1\leq i \leq |\mathcal{\mathcal{S}}|)$, where the term $t_{ij}$ is semantically close to $s_i$ and far from other seeds $s_j$ $(\forall j \neq i)$.
\label{def:problem}
\end{definition}

In other words, each seed $s_i$ represents a semantic category $c_i$, and the task is to find a set of terms $\mathcal{T}_i$ that discriminatively belong to the category $c_i$ $(1\leq i \leq |\mathcal{\mathcal{S}}|)$.

\section{Framework}
In this section, we first review various types of contexts utilized by previous studies for topic discovery. Then, we present our framework, \textsc{\model}, that iteratively ensembles these types of signals.

\begin{figure*}
\centering
\includegraphics[width=0.95\linewidth]{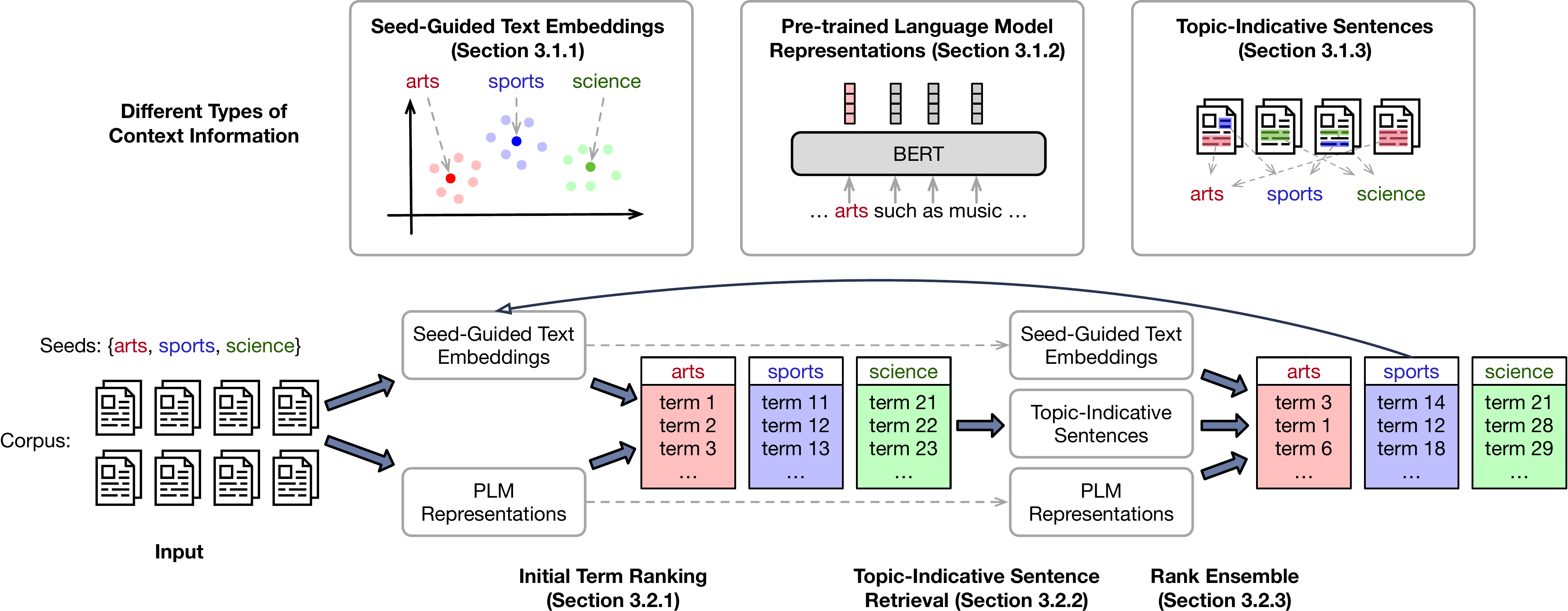}
\caption{Overview of the \textsc{\model} framework.} 
\vspace{-0.5em}
\label{fig:framework}
\end{figure*}

\subsection{Types of Context Information}
Previous studies on topic discovery, either unsupervised, seed-guided, or supervised, propose to leverage different types of information such as skip-gram embeddings \cite{meng2020discriminative,dieng2020topic,harandizadeh2022keyword,xun2017collaboratively,meng2020hierarchical}, pre-trained language model representations \cite{meng2022topic,sia2020tired,thompson2020topic,bianchi2021pre,zhang2022seed}, and topic-indicative documents \cite{mcauliffe2007supervised,lacoste2008disclda,lee2022taxocom}. We now introduce three major types of information sources, which are illustrated in Figure \ref{fig:framework}, and how we propose to use them in \textsc{\model}.

\subsubsection{Seed-Guided Text Embeddings.} 
\label{sec:emb}
Previous embedding-based topic models \cite{dieng2020topic,harandizadeh2022keyword,xun2017collaboratively} propose to incorporate word embeddings to make up for the representation deficiency of the ``bag-of-words'' generation assumption in LDA. The intuition of text embedding learning is based on the hypothesis that semantically similar terms share similar contexts. In unsupervised topic discovery, the contexts of a term may refer to its skip-grams \cite{mikolov2013distributed} and the documents it appears in \cite{le2014distributed,tang2015pte}. In our task of seed-guided topic discovery, we can further leverage seeds in the embedding learning process by viewing the category that a term belongs to as its context.
To facilitate this goal, in \textsc{\model}, we follow \cite{meng2020discriminative,zhang2022seed} and unify the three types of contexts into one objective. To be specific, we aim to maximize the likelihood of observing a term's skip-gram, document, and category contexts given that term. Formally, the embedding learning objective is:
\begin{equation}
\small
\begin{split}
    \mathcal{J}_{\rm Emb} = & \underbrace{\log \prod_{d\in\mathcal{D}}\prod_{w_i \in d}\ \prod_{w_j\in \mathcal{C}(w_i)} p(w_j|w_i)}_{\rm skip-gram} \\
    & + \ \underbrace{\log \prod_{d\in \mathcal{D}}\prod_{w\in d} p(d|w)}_{\rm document} \ + \ \underbrace{\log \prod_{1\leq i \leq |\mathcal{\mathcal{S}}|}\prod_{w\in \mathcal{T}_i} p(c_i|w)}_{\rm category}.
\end{split}
\label{eqn:emb}
\end{equation}
Here, $\mathcal{C}(w_i)$ is the set of terms in $w_i$'s skip-gram window. For example, given a text sequence $w_1w_2...w_X$, we have $\mathcal{C}(w_i) = \{w_j|i-x \leq j \leq i+x, \ j \neq i\}$, where $x$ is the skip-gram window size. $\mathcal{T}_i$, as mentioned in Definition \ref{def:problem}, is the set of terms related to the seed $s_i$ (i.e., belong to the semantic category $c_i$). We adopt an iterative framework to gradually expand $\mathcal{T}_i$. At the very beginning, $\mathcal{T}_i=\{s_i\}$ (i.e., the seed initially belongs to its corresponding topic). After each iteration, terms close to the category $c_i$ in the embedding space will be added to $\mathcal{T}_i$.

There are various ways to define each likelihood in Eq. (\ref{eqn:emb}). Following previous studies on topic modeling \cite{meng2020discriminative,batmanghelich2016nonparametric,li2016integrating,jameel2019word}, we adopt the von Mises-Fisher (vMF) distribution.
\begin{equation}
\small
    p(w_j|w_i) = \frac{\exp(\kappa_{w_i} \cos(\bmu_{w_i}, \bmv_{w_j}))}{\sum_{w'} \exp(\kappa_{w_i} \cos(\bmu_{w_i}, \bmv_{w'}))} \simeq {\rm vMF}(\bmv_{w_j}|\bmu_{w_i}, \kappa_{w_i}),
\label{eqn:vmf}
\end{equation}
where $\kappa_{w_i} \geq 0$ is the concentration parameter, indicating the semantic specificity of $w_i$; $\bmu_{w_i}$ and $\bmv_{w_j}$ are the embeddings of $w_i$ and $w_j$, respectively. The vMF distribution can be viewed as an analogue of the Gaussian distribution on a sphere. In Eq. (\ref{eqn:vmf}), the distribution concentrates around the mean direction $\bmu_{w_i}$, and is more concentrated if $\kappa_{w_i}$ is larger (i.e., $w_i$ is a more specific term). Similar to Eq. (\ref{eqn:vmf}), the other two likelihood terms in Eq. (\ref{eqn:emb}) can be defined as
\begin{equation}
\small
\begin{split}
p(d|w) &= \frac{\exp(\kappa_w \cos(\bmu_w, \bmv_d))}{\sum_{d'} \exp(\kappa_w \cos(\bmu_w, \bmv_{d'}))} \simeq {\rm vMF}(\bmv_d|\bmu_w, \kappa_w), \\
p(c_i|w) &= \frac{\exp(\kappa_w \cos(\bmu_w, \bmv_{c_i}))}{\sum_{c'} \exp(\kappa_w \cos(\bmu_w, \bmv_{c'}))} \simeq {\rm vMF}(\bmv_{c_i}|\bmu_w, \kappa_w),
\end{split}
\end{equation}
where $\bmv_d$ and $\bmv_{c_i}$ are the embedding vectors of document $d$ and category $c_i$, respectively.

To summarize, the seed-guided text embedding learning process is cast as the following optimization problem:
\begin{equation}
\small
\max \mathcal{J}_{\rm Emb} \ \ {\rm s.t.} \  ||\bmu_w||=||\bmv_w||=||\bmv_d||=||\bmv_c||=1, \  \kappa_w \geq 0.
\label{eqn:optimization}
\end{equation}
We follow the optimization process of \cite{meng2020discriminative} to optimize Eq. (\ref{eqn:optimization}).

After embedding learning, for each term $w$, we obtain two vectors $\bmu_w$ and $\bmv_w$, which, as they do in previous studies \cite{mikolov2013distributed,tang2015pte,meng2020discriminative}, carry the semantics of $w$ when it is viewed as a center term and a context term, respectively. Given a term $w$ and a seed $s_i$, we calculate the cosine similarity between their learned embeddings as the first criterion of their semantic proximity, which will later be used in topic discovery.
\begin{equation}
\small
{\rm sim}_{\rm Emb}(w, s_i) = \cos(\bmu_w, \bmu_{s_i}).
\label{eqn:emb_sim}
\end{equation}

\subsubsection{Pre-trained Language Model Representations}
\label{sec:plm}
Recently, PLMs such as BERT \cite{devlin2019bert} have achieved great success in a wide spectrum of text mining tasks. The Transformer architecture \cite{vaswani2017attention} used in many PLMs is capable of capturing long-range and high-order context signals. Moreover, the generic knowledge learned by PLMs from web-scale corpora (e.g., Wikipedia) can complement the information one can get from the input corpus. To utilize such signals in topic discovery, for each term appearing in the input corpus, we employ a PLM to derive its representation.

Suppose a term $w$ appears $M$ times in the corpus $\mathcal{D}$. For each of its mentions $w^i$ $(1 \leq i \leq M)$, we feed the sentence containing this mention into a PLM. Note that $w^i$ may be segmented into multiple word pieces $w^i_1, w^i_2, ..., w^i_L$ according to the PLM tokenizer \cite{schuster2012japanese,sennrich2016neural}, and each word piece $w^i_j$ will have an output representation vector ${\bf PLM}(w^i_j)$ after PLM encoding. Following previous studies on topic discovery \cite{sia2020tired,thompson2020topic}, we take the average of these word piece representations as the representation of the mention.
\begin{equation}
\small
    {\bf PLM}(w^i) = \frac{1}{L} \sum_{j=1}^L {\bf PLM}(w^i_j).
\end{equation}
The mention representation is contextualized given the sentence it appears in. To get the corpus-level semantics of a term, we average the representations of all its mentions.
\begin{equation}
\small
    \bmh_w = \frac{1}{M} \sum_{i=1}^{M} {\bf PLM}(w^i).
\label{eqn:plm}
\end{equation}

In this way, for each $w$, we obtain a vector $\bmh_w$ whose dimension is given by the adopted PLM. For example, if we use $\rm BERT_{Base}$ \cite{devlin2019bert}, then $\bmh_w \in \mathbb{R}^{768}$. Given a term $w$ and a seed $s_i$, we calculate the cosine similarity between their PLM-based representations as our second criterion of their semantic proximity.
\begin{equation}
\small
    {\rm sim}_{\rm PLM}(w, s_i) = \cos(\bmh_w, \bmh_{s_i}).
\label{eqn:plm_sim}
\end{equation}

\subsubsection{Topic-Indicative Context}
Although seed-guided embedding learning and PLM encoding are both powerful tools to represent each term based on its contexts, neither of them considers whether the utilized context information is topic-indicative or not. To be specific, the PLM-based representation $\bmh_w$ is unaware of the seed space $\mathcal{S}$ (in other words, no matter what the seeds $\mathcal{S}=\{s_1,...,s_{|\mathcal{\mathcal{S}}|}\}$ are, if the corpus $\mathcal{D}$ is fixed, then the same PLM will always generate the same representation vector $\bmh_w$ for $w$); the embeddings $\bmu_w$ and $\bmv_w$ always take the skip-gram $\mathcal{C}(w)$ (i.e., $\pm 1, ..., \pm x$ terms) and the document $d$ containing $w$ as contexts during learning, regardless of whether such information is relevant to a certain seed/topic. To alleviate this gap, we propose to use topic-indicative context to derive the correlation between a term $w$ and a seed $s_i$.

For each seed $s_i \in \mathcal{S}$, we assume it has a set of topic-indicative sentences $\Theta_i=\{\theta_{i1},...,\theta_{i|\Theta_i|}\}$. (Initially, $\Theta_i$ is not given as input. We will discuss how to obtain and iteratively update $\Theta_i$ in Section \ref{sec:seg_retr}.) The reason that we consider sentences instead of documents here is because a document is more likely to cover multiple topics. Motivated by \cite{tao2016multi,lee2022taxocom}, we calculate the semantic closeness between $w$ and $\Theta_i$ according to the following two criteria:
(1) \textit{Popularity}: a term close to $\Theta_i$ should appear frequently in the sentences in $\Theta_i$. Formally, ${\rm pop}(w, \Theta_i) = \log(1+{\rm tf}(w, \Theta_i))$, where $\rm tf(\cdot, \cdot)$ denotes term frequency and ${\rm tf}(w, \Theta_i) = \sum_{j=1}^{|\Theta_i|}{\rm tf}(w, \theta_{ij})$.
(2) \textit{Distinctiveness}: a term close to $\Theta_i$ should be much more relevant to the sentences in $\Theta_i$ than it is to the sentences indicating other topics. This can be characterized by the formula:
${\rm dist}(w, \Theta_i) = \frac{\exp({\rm BM25}(w, \Theta_i))}{1+\sum_{i'=1}^{|\mathcal{S}|}\exp({\rm BM25}(w, \Theta_{i'}))}$, where $\rm BM25(\cdot, \cdot)$ denotes the BM25 relevance function \cite{robertson1994some}.

To jointly consider popularity and distinctiveness, the similarity between a term $w$ and a category $c_i$ based on topic-indicative sentences is defined as follows.
\begin{equation}
\small
{\rm sim}_{\rm Sntn}(w, c_i) = {\rm pop}(w, \Theta_i)^{\alpha} \cdot {\rm dist}(w, \Theta_i)^{1-\alpha},
\label{eqn:seg_sim}
\end{equation}
where $0<\alpha<1$ is a hyperparameter.

\subsection{The Iterative \textsc{\model} Framework}
We lay out our framework in Figure \ref{fig:framework}. It has three major modules: initial term ranking, topic-indicative sentence retrieval, and rank ensemble. We now introduce these modules in detail.

\subsubsection{Initial Term Ranking}
Initially, we only have the seed $s_i$ for each semantic category $c_i$, and the topic-indicative sentences $\Theta_i$ have not been derived yet. Therefore, we first use seed-guided text embeddings (derived in Section \ref{sec:emb}) and PLM-based representations (derived in Section \ref{sec:plm}) to find terms that are relevant to each category. To be specific, for each category $c_i$, we calculate the following score for each term $w$.
\begin{equation}
\small
    {\rm score}_{\rm Ini}(w, c_i) = {\rm sim}_{\rm Emb}(w, s_i) \cdot {\rm sim}_{\rm PLM}(w, s_i),
\label{eqn:score1}
\end{equation}
where ${\rm sim}_{\rm Emb}(\cdot, \cdot)$ and ${\rm sim}_{\rm PLM}(\cdot, \cdot)$ are given in Eqs. (\ref{eqn:emb_sim}) and (\ref{eqn:plm_sim}), respectively.
As mentioned in Section \ref{sec:emb}, the set of topic-related terms $\mathcal{T}_i$ is expanded and updated iteratively. In later iterations, when $\mathcal{T}_i$ is more than just $\{s_i\}$, Eq. (\ref{eqn:score1}) can be generalized to
\begin{equation}
\small
{\rm score}_{\rm Ini}(w, c_i) = \sum_{t_{ij} \in \mathcal{T}_i} {\rm sim}_{\rm Emb}(w, t_{ij}) \cdot \sum_{t_{ij} \in \mathcal{T}_i} {\rm sim}_{\rm PLM}(w, t_{ij}).
\label{eqn:score2}
\end{equation}
For each category $c_i$, we find top-$\tau$ terms according to ${\rm score}_{\rm Ini}(w,$ $c_i)$ to update the topic-indicative term set $\mathcal{T}_i$.

\newlength{\textfloatsepsave} 
\setlength{\textfloatsepsave}{\textfloatsep}
\setlength{\textfloatsep}{0pt}
\begin{algorithm}[t]
\caption{\textsc{\model}}
\small
\label{alg:seedtopic}
\KwIn{A corpus $\mathcal{D}$; \  a set of seeds $\mathcal{S}=\{s_1,...,s_{|\mathcal{\mathcal{S}}|}\}$.}
\KwOut{A set of terms $\mathcal{T}_i=\{t_{i1},t_{i2},...,t_{i|\mathcal{T}_i|}\}$ appearing in $\mathcal{D}$ for each seed $s_i$.}
$\mathcal{T}_i = \{s_i\}$\;
$\bmh_w \gets$ Eq. (\ref{eqn:plm})\;
\For{${\rm iter} \gets 1$ to $N$}
{
    Learn seed-guided text embedding $\bmu_w$ by optimizing Eq. (\ref{eqn:optimization})\;
    \textcolor{myblue}{// Initial Term Ranking\;}
    ${\rm score}_{\rm Ini}(w, c_i) \gets$ Eq. (\ref{eqn:score2})\;
    $\mathcal{T}_i \gets$ top-ranked terms according to ${\rm score}_{\rm Ini}(w, c_i)$\;
    \textcolor{myblue}{// Topic-Indicative Sentence Retrieval\;}
    ${\rm count}(\theta, c_i) \gets$ Eq. (\ref{eqn:score3})\;
    $\Theta_i^{A} \gets$ top-ranked sentences according to Eq. (\ref{eqn:anchor})\;
    $\Theta_i^{N} \gets \emptyset$ \;
    \For{$\theta_{ij} \in \Theta_i^{A}$}
    {
        \For{$k \gets 1$ to $y$}
        {
            Denote the $+k$ sentence of $\theta_{ij}$ as $\theta_{ij}^{+k}$\;
            \If{$\forall i' \neq i$, ${\rm count}(\theta_{ij}^{+k}, c_{i'}) = 0$}
            {
                $\Theta_i^{N} \gets \Theta_i^{N} \cup \{\theta_{ij}^{+k}\}$\;
            }
            \uElse
            {
                \textbf{break}\;
            }
        }
        \For{$k \gets 1$ to $y$}
        {
            Denote the $-k$ sentence of $\theta_{ij}$ as $\theta_{ij}^{-k}$\;
            \If{$\forall i' \neq i$, ${\rm count}(\theta_{ij}^{-k}, c_{i'}) = 0$}
            {
                $\Theta_i^{N} \gets \Theta_i^{N} \cup \{\theta_{ij}^{-k}\}$\;
            }
            \uElse
            {
                \textbf{break}\;
            }
        }
    }
    $\Theta_i \gets \Theta_i^{A} \cup \Theta_i^{N}$\;
    \textcolor{myblue}{// Rank Ensemble\;}
    ${\rm score}_{\rm All}(w, c_i) \gets$ Eq. (\ref{eqn:score4})\;
    ${\rm MRR}(w|c_i) \gets$ Eq. (\ref{eqn:mrr})\;
    $\mathcal{T}_i \gets$ Eq. (\ref{eqn:update})\;
}
$\mathcal{T}_i \gets \mathcal{T}_i \backslash \{s_i\}$\;
Return $\mathcal{T}_1, ..., \mathcal{T}_{|\mathcal{S}|}$\;
\end{algorithm}
\setlength{\textfloatsep}{\textfloatsepsave}

\subsubsection{Topic-Indicative Sentence Retrieval}
\label{sec:seg_retr}
Based on the set of updated topic-indicative terms $\mathcal{T}_i$, we now retrieve the set of topic-indicative sentences $\Theta_i$ from the input corpus so that we can calculate Eq. (\ref{eqn:seg_sim}). The retrieval process is inspired by two assumptions: (1) The sentences containing many topic-indicative terms from one category and do not contain any topic-indicative term from other categories should be topic-indicative sentences. We call such sentences ``\textit{anchor}'' sentences. (2) The ``\textit{neighbor}'' sentences of topic-indicative ``anchor'' sentences should also be viewed as topic-indicative if they do not contain topic-indicative terms from other categories. 

According to Assumption (1), we first retrieve ``anchor'' sentences by counting the number of topic-indicative terms appearing in each sentence. Formally, given a category $c_i$, for each sentence $\theta$ in $\mathcal{D}$, we calculate
\begin{equation}
\small
{\rm count}(\theta, c_i) = \sum_{w \in \mathcal{T}_i} {\rm tf}(w, \theta).
\label{eqn:score3}
\end{equation}
Because category-indicative ``anchor'' sentences should have a high count with $c_i$ and a count of $0$ with any other category, we rank the sentences using the following criterion.
\begin{equation}
\small
\max_{\theta \in \mathcal{D}} {\rm count}(\theta, c_i), \ \ \  \text{where} \ {\rm count}(\theta, c_j) = 0 \ \ \ (\forall j \neq i).
\label{eqn:anchor}
\end{equation}
We use $\Theta_i^{A}=\{\theta_{i1},...,\theta_{i|\Theta_i^{A}|}\}$ to denote the set of selected ``anchor'' sentences for $c_i$.

Then, according to Assumption (2), we find ``neighbor'' sentences for each ``anchor'' sentence $\theta_{ij}$. To be specific, given an ``anchor'' sentence, we check its $\pm 1, \pm 2, ..., \pm y$ sentences in the document (if they exist). If the $+k$ (resp., $-k$) sentence contains topic-indicative terms from other categories, we view it as not topic-indicative, and we do not further check the $+(k+1),...,+y$ (resp., $-(k+1),...,-y$) sentences because they may have diverged to other topics. Otherwise, we add the $+k$ (resp., $-k$) sentence into the set of topic-indicative ``neighbor'' sentences $\Theta_i^{N}$. A more formal description of this process can be found in Lines 13--24 in Algorithm \ref{alg:seedtopic}. 

Finally, the set of retrieved topic-indicative sentences is the union of ``anchor'' sentences and ``neighbor'' sentences (i.e., $\Theta_i = \Theta_i^{A} \cup \Theta_i^{N}$).

\subsubsection{Ensemble of Multiple Types of Contexts}
After obtaining topic-indicative context $\Theta_i$ of each category $c_i$, we can now calculate ${\rm sim}_{\rm Sntn}(w, c_i)$ in Eq. (\ref{eqn:seg_sim}). Then, we have a score measuring the semantic proximity between a term $w$ and a category $c_i$ by jointly considering all three types of contexts.
\begin{equation}
\small
\begin{split}
& {\rm score}_{\rm All}(w, c_i) = \\
& \sum_{t_{ij} \in \mathcal{T}_i} {\rm sim}_{\rm Emb}(w, t_{ij}) \cdot \sum_{t_{ij} \in \mathcal{T}_i} {\rm sim}_{\rm PLM}(w, t_{ij}) \cdot {\rm sim}_{\rm Sntn}(w, c_i).
\end{split}
\label{eqn:score4}
\end{equation}
By ranking all terms in a descending order of ${\rm score}_{\rm All}(w, c_i)$, we get a ranking list where each term $w$ has a rank position $r_{\rm All}(w|c_i)$. Besides, instead of incorporating topic-indicative context into ranking, we can consider seed-guided text embeddings alone or PLM-based representations alone. By ranking terms in a descending order of $\sum_{t_{ij} \in \mathcal{T}_i} {\rm sim}_{\rm Emb}(w, t_{ij})$ and $ \sum_{t_{ij} \in \mathcal{T}_i} {\rm sim}_{\rm PLM}(w, t_{ij})$, each term $w$ will have two more rank positions $r_{\rm Emb}(w|c_i)$ and $r_{\rm PLM}(w|c_i)$, respectively. Based on the three rank positions, we perform rank ensemble by calculating the mean reciprocal rank (MRR).
\begin{equation}
\small
    {\rm MRR}(w|c_i) = \frac{1}{3}\bigg(\frac{1}{r_{\rm All}(w|c_i)} + \frac{1}{r_{\rm Emb}(w|c_i)} + \frac{1}{r_{\rm PLM}(w|c_i)}\bigg).
\label{eqn:mrr}
\end{equation}
In practice, instead of ranking all terms in the vocabulary, we only check the top-$\rho$ terms in each ranking list. If a term $w$ is not among the top-$\rho$ (e.g., $r_{\rm All}(w|c_i) > \rho$), we simply set its reciprocal rank to be 0 (e.g., $\frac{1}{r_{\rm All}(w|c_i)} = 0$).
Finally, we update $\mathcal{T}_i$ with the terms whose MRR score exceeds a certain threshold $\eta$.
\begin{equation}
\small
    \mathcal{T}_i = \{w\ |\ {\rm MRR}(w|c_i) \geq \eta\}, \ \ (1\leq i \leq |\mathcal{S}|).
\label{eqn:update}
\end{equation}
The updated term sets $\mathcal{T}_1, ..., \mathcal{T}_{|\mathcal{S}|}$ are then fed into the next iteration of \textsc{\model}. 

We iterate the process of initial term ranking, topic-indicative sentence retrieval, and rank ensemble for $N$ iterations.
The entire \textsc{\model} framework is summarized in Algorithm \ref{alg:seedtopic}.

\newcommand{\bs}[1]{\boldsymbol{#1}}
\renewcommand{\textrightarrow}{$\rightarrow$}
\newcommand{\comments}[1]{\textcolor{red}{#1}}
\newcommand{\wrong}[1]{#1 ($\times$)}
\newcommand{\wrongnspc}[1]{#1($\times$)}

\section{Experiments}

\setlength{\tabcolsep}{3pt}
\begin{table}[t]
\small
\caption{Dataset Statistics.}
\vspace{-1em}
\scalebox{0.925}{
\begin{tabular}{c|c|c|c|c}
\toprule
Dataset & \multicolumn{2}{c|}{\textbf{NYT}} & \multicolumn{2}{c}{\textbf{Yelp}} \\
\midrule
Dimension & Topic & Location & Food & Sentiment \\
\#Docs & 31,997 & 31,997 & 29,280 & 29,280 \\
\#Seeds & 9 & 10 & 8 & 2 \\
\midrule
Seeds & \makecell{arts, technology, \\ health, education, \\ sports, science, \\ business, politics, \\ real estate} & \makecell{united states, \\ iraq, britain, \\ japan, canada, \\ china, france, \\ italy, russia, \\ germany} & \makecell{steak, seafood, \\ pizza, desserts, \\ salad, noodles, \\ sushi, burgers} & good, bad \\
\bottomrule
\end{tabular}
}
\vspace{-0.5em}
\label{tab:data}
\end{table}

\subsection{Setup}
\subsubsection{Datasets}
Following \cite{meng2020discriminative}, we conduct experiments on two datasets from different domains.
\begin{itemize}[leftmargin=*]
\item \textbf{NYT}\footnote{\url{https://catalog.ldc.upenn.edu/LDC2008T19}} is a collection of news articles written and published by the New York Times. It has two sets of seeds along the \textit{topic} and \textit{location} dimensions, respectively.
\item \textbf{Yelp}\footnote{\url{https://www.yelp.com/dataset/challenge}} is a corpus of restaurant reviews released by the Yelp Dataset Challenge. It has two sets of seeds along the \textit{food} and \textit{sentiment} dimensions, respectively.
\end{itemize}
For both datasets, we use AutoPhrase \cite{shang2018automated} to perform phrase chunking. Following \cite{sia2020tired}, we adopt a 60-40 train-test split for both datasets. The training set is used as the input corpus $\mathcal{D}$, and the testing set is used to calculate the topic coherence metric (see evaluation metrics for details). Dataset statistics are summarized in Table \ref{tab:data}.

\begin{table*}[t]
\small
\centering
\caption{NPMI, P@20, and NDCG@20 scores of compared algorithms. NPMI measures topic coherence; P@20 and NDCG@20 measure term accuracy.}
\vspace{-0.5em}
\scalebox{0.925}{
	\begin{tabular}{c|p{0.95cm}<{\centering} p{0.95cm}<{\centering} c|p{0.95cm}<{\centering} p{0.95cm}<{\centering} c|p{0.95cm}<{\centering} p{0.95cm}<{\centering} c|p{0.95cm}<{\centering} p{0.95cm}<{\centering} c}
		\toprule
		\multirow{2}{*}{Method} &
		\multicolumn{3}{c|}{\textbf{NYT}-Topic} & \multicolumn{3}{c|}{\textbf{NYT}-Location} & \multicolumn{3}{c|}{\textbf{Yelp}-Food} & \multicolumn{3}{c}{\textbf{Yelp}-Sentiment} \\
		& NPMI & P@20 & NDCG@20 & NPMI & P@20 & NDCG@20 & NPMI & P@20 & NDCG@20 & NPMI & P@20 & NDCG@20 \\
		\midrule
		SeededLDA \cite{jagarlamudi2012incorporating} & 0.0841 & 0.2389 & 0.2979 & 0.0814 & 0.1050 & 0.1873 & 0.0504 & 0.1200 & 0.2132 & 0.0499 & 0.1700 & 0.2410 \\
		Anchored CorEx \cite{gallagher2017anchored}   & 0.1325 & 0.2922 & 0.3627 & 0.1283 & 0.2040 & 0.3003 & 0.1204 & 0.3725 & 0.4531 & 0.0627 & 0.1200 & 0.1997 \\
		KeyETM \cite{harandizadeh2022keyword}         & 0.1254 & 0.1589 & 0.2342 & 0.1146 & 0.0700 & 0.1676 & 0.0578 & 0.1788 & 0.2940 & 0.0327 & 0.4250 & 0.4994 \\
		CatE \cite{meng2020discriminative}            & 0.1941 & 0.8067 & 0.8306 & 0.2165 & 0.7480 & 0.7840 & \textbf{0.2058} & 0.6812 & 0.7312 & \textbf{0.1509} & 0.7150 & 0.7713 \\
		\textsc{\model} & \textbf{0.1947} & \textbf{0.9456} & \textbf{0.9573} & \textbf{0.2176} & \textbf{0.8360} & \textbf{0.8709} & 0.2018 & \textbf{0.7912} & \textbf{0.8379} & 0.0922 & \textbf{0.9750} & \textbf{0.9811} \\
		\bottomrule
	\end{tabular}
}
\label{tab:performance}
\vspace{-0.5em}
\end{table*}

\subsubsection{Compared Methods}
We compare our \textsc{\model} with the following baselines including seed-guided topic modeling methods and seed-guided embedding learning methods.
\begin{itemize}[leftmargin=*]
\item \textbf{SeededLDA \cite{jagarlamudi2012incorporating}} is a seed-guided topic modeling method. It modifies the generative process of LDA by biasing each topic to generate more seeds and by biasing each document to select topics relevant to the seeds appearing in the document.
\item \textbf{Anchored CorEx \cite{gallagher2017anchored}} is a seed-guided topic modeling method. It does not rely on generative assumptions. Instead, it leverages seeds by balancing between compressing the input corpus and preserving seed-related information.
\item \textbf{KeyETM \cite{harandizadeh2022keyword}} is an embedding-based topic model (ETM) assisted by keyword seeds. It modifies the objective of ETM \cite{dieng2020topic} to utilize seeds in the form of topic-level priors over the vocabulary.
\item \textbf{CatE \cite{meng2020discriminative}} is a seed-guided embedding learning method for discriminative topic mining. It jointly learns term embedding and specificity from the input corpus. Terms are then selected based on both embedding similarity with the seeds and specificity.
\end{itemize}

For unsupervised topic discovery approaches (e.g., BERTopic \cite{grootendorst2022bertopic} and TopClus \cite{meng2022topic}), it is difficult to match their generated topics to the given seeds, so we cannot calculate term accuracy-based metrics (see evaluation metrics for details) for their output, and hence we do include them into comparison.

\subsubsection{Evaluation Metrics}
Given the top-$|\mathcal{T}_i|$ discovered terms under each seed ($|\mathcal{T}_i|=20$ in our experiments), we evaluate the results based on two different criteria: \textit{topic coherence} and \textit{term accuracy}.
\begin{itemize}[leftmargin=*]
\item \textbf{NPMI \cite{lau2014machine}} is a widely adopted metric in topic modeling to measure \textit{topic coherence} inside each topic. It is defined as the average normalized pointwise mutual information of each pair of terms in $\mathcal{T}_i$.
\begin{equation}
\small
    {\rm NPMI} = \frac{1}{|\mathcal{S}|}\sum_{i=1}^{|\mathcal{S}|} \frac{1}{\binom{|\mathcal{T}_i|}{2}}\sum_{t_{ij}, t_{ik} \in \mathcal{T}_i} \frac{\log\frac{P(t_{ij}, t_{ik})}{P(t_{ij})P(t_{ik})}}{-\log P(t_{ij}, t_{ik})},
    \label{eqn:npmi}
\end{equation}
where $P(t_{ij}, t_{ik})$ is the probability that $t_{ij}$ and $t_{ik}$ co-occur in a document; $P(t_{ij})$ is the probability that $t_{ij}$ occurs in a document.
\item \textbf{P@$\bm k$ (also called MACC) \cite{meng2020discriminative}} is a metric for \textit{term accuracy}. It measures the proportion of retrieved terms $t_{ij}$ that actually belong to the semantic category $c_i$.
\begin{equation}
\small
    {\rm P@}k = \frac{1}{|\mathcal{S}|}\sum_{i=1}^{|\mathcal{S}|}\frac{1}{|\mathcal{T}_i|}\sum_{t_{ij} \in \mathcal{T}_i} {\bf 1}(t_{ij} \in c_i),
    \label{eqn:prec}
\end{equation}
where ${\bf 1}(t_{ij} \in c_i)$ is the indicator function of whether $t_{ij}$ belongs to $c_i$ (i.e., whether $t_{ij}$ is discriminatively relevant to the seed $s_i$). This relies on human judgment, so we invite five annotators to perform independent annotation. The reported P@$k$ score is the average P@$k$ of the five annotators. A high inter-annotator agreement is observed, with Fleiss' kappa \cite{fleiss1971measuring} being 0.896, 0.928, 0.800, and 0.909 on NYT-Topic, NYT-Location, Yelp-Food, and Yelp-Sentiment, respectively. As mentioned above, we set $|\mathcal{T}_i|=20$ in our experiments, so we report P@20.
\item \textbf{NDCG@$\bm k$} is another metric for \textit{term accuracy}. It gives higher weights to higher-ranked terms by applying a logarithmic discount.
\begin{equation}
\small
\begin{split}
    {\rm DCG}_i@k = \sum_{j=1}^{|\mathcal{T}_i|} \frac{{\bf 1}(t_{ij} \in c_i)}{\log(j+1)}, & \ \ \ \ {\rm IDCG}@k = \sum_{j=1}^{|\mathcal{T}_i|} \frac{1}{\log(j+1)}, \\
    {\rm NDCG@}k = & \ \frac{1}{|\mathcal{S}|}\sum_{i=1}^{|\mathcal{S}|} \frac{{\rm DCG}_i@k}{{\rm IDCG}@k}.
\end{split}
\end{equation}
Following the case of P@$k$, we calculate NDCG@$k$ based on human annotations, and we report NDCG@20.
\end{itemize}

\subsubsection{Hyperparameters and Implementation} The hyperparameter settings of \textsc{\model} are as follows. In seed-guided embedding learning, the context window size $x=5$; the embedding dimension is 100. In PLM encoding, we use $\rm BERT_{Base}$ \cite{devlin2019bert} as the PLM. When computing ${\rm sim}_{\rm Sntn}(w, c_i)$, we set $\alpha=0.2$. In initial term ranking, we select $\tau=20$ terms for each seed. In topic-indicative sentence retrieval, we retrieve $|\Theta_i^{A}|=500$ ``anchor'' sentences; the ``neighbor'' sentence window size $y=4$. In rank ensemble, there are $\rho=20$ terms in each ranking list; the MRR threshold $\eta=0.1$. We run \textsc{\model} for $N=4$ iterations.

The code, datasets, and annotation results are available at \textcolor{myblue}{\url{https://github.com/yzhan238/SeedTopicMine}}.

\begin{table*}[t]
\small
\centering
\caption{Top-5 terms retrieved by different algorithms. $\times$: At least 3 of the 5 annotators judge the term as irrelevant to the seed.}
\vspace{-0.5em}
\scalebox{0.925}{
	\begin{tabular}{c|cc|cc|cc|cc}
		\toprule
		\multirow{2}{*}{Method} &		\multicolumn{2}{c|}{\textbf{NYT}-Topic} &
		\multicolumn{2}{c|}{\textbf{NYT}-Location} &
		\multicolumn{2}{c|}{\textbf{Yelp}-Food} &
		\multicolumn{2}{c}{\textbf{Yelp}-Sentiment} \\
		& health & business & france & canada & sushi & desserts & good & bad \\
		\midrule
		
		\multirow{5}{*}{SeededLDA} & \wrong{said} & \wrong{said} & \wrong{said}  & \wrong{new}  & roll & \wrong{food} & \wrong{place} & \wrong{food} \\
		& \wrong{dr} & \wrong{percent} & \wrong{new} & \wrong{city} & \wrong{good} & \wrong{us} & \wrong{food} & \wrong{service} \\
		& \wrong{new} & company & \wrong{state} & \wrong{said} & \wrong{place} & \wrong{order} & great & \wrong{us} \\
		& \wrong{would} & \wrong{year} & \wrong{would} & \wrong{building} & \wrong{food} & \wrong{service} & \wrong{like} & \wrong{order} \\
		& hospital & \wrong{billion} & \wrong{dr} & \wrong{mr} & rolls & \wrong{time} & \wrong{service} & \wrong{time}\\
		\midrule
		
		\multirow{5}{*}{\begin{tabular}[c]{@{}c@{}} Anchored \\ CorEx \end{tabular}} & \wrong{case} & employees & \wrong{school} & \wrong{market} & rolls & \wrong{also} & \wrong{definitely} & \wrong{one} \\
		& \wrong{court} & advertising & \wrong{students} & \wrong{percent} & roll & \wrong{really} & \wrong{prices} & \wrong{would} \\
		& patients & \wrong{media} & \wrong{children} & \wrong{companies} & sashimi & \wrong{well} & \wrong{strip} & \wrong{like} \\
		& \wrong{cases} & businessmen & \wrong{education} & \wrong{billion} & \wrong{fish} & \wrong{good} & \wrong{selection} & \wrong{could} \\
		& \wrong{lawyer} & commerce & \wrong{schools} & \wrong{investors} & tempura & \wrong{try} & \wrong{value} & \wrong{us} \\
		\midrule

		\multirow{5}{*}{KeyETM} & \wrong{team} & \wrong{percent} & \wrong{city} & \wrong{people} & sashimi & \wrong{food} & great & \wrong{food}   \\
		& \wrong{game} & \wrong{japan} & \wrong{state} & \wrong{year} & rolls & \wrong{great} & delicious & \wrong{place}  \\
		& \wrong{players} & \wrong{year} & \wrong{york} & \wrong{china} & roll & \wrong{place} & amazing & \wrong{service}   \\
		& \wrong{games} & \wrong{japanese} & \wrong{school} & \wrong{years} & \wrong{fish} & \wrong{good} & excellent & \wrong{time} \\
		& \wrong{play} & economy & \wrong{program} & \wrong{time} & japanese & \wrong{service} & tasty & \wrong{restaurant}  \\
		\midrule
		
		\multirow{5}{*}{CatE} & public health & \wrong{diversifying} & french & alberta & \wrong{freshest fish} & \wrong{delicacies} & tasty & unforgivable \\
		& health care & \wrong{clients} & corsica & british columbia & sashimi & sundaes & delicious & frustrating \\
		& medical & corporate & \wrong{spain} & ontario & nigiri & \wrong{savoury} & yummy & horrible \\
		& hospitals & investment banking & \wrong{belgium} & manitoba & ayce sushi & pastries & \wrong{chilaquiles} & irritating \\
		& doctors & executives & \wrong{de} & canadian & rolls & custards & \wrong{also} & rude \\
		\midrule
		
		\multirow{5}{*}{\textsc{\model}} & medical & companies & french & canadian & maki rolls & cheesecakes & great & terrible \\
		& hospitals & businesses & paris & quebec & sashimi & croissants & excellent & horrible \\
		& hospital & corporations & \wrong{philippe} & montreal & ayce sushi & pastries & fantastic & awful \\
		& {public health} & firms & {french state} & toronto & revolving sushi & \wrong{breads} & delicious & lousy \\
		& patients & corporate & frenchman & ottawa & nigiri & cheesecake & amazing & shitty \\

		\bottomrule
	\end{tabular}
}
\label{tab:case}
\vspace{-0.5em}
\end{table*}

\subsection{Performance Comparison}
Table \ref{tab:performance} shows the NPMI, P@20, and NDCG@20 scores of compared algorithms on the two datasets.
We can observe that: (1) On NYT, \textsc{\model} consistently achieves the best performance in terms of all metrics. Among all the baselines, CatE is the most effective one, significantly outperforming ``bag-of-words''-based topic models such as SeededLDA and Anchored CorEx. However, since CatE only uses one type of context information (i.e., skip-gram embeddings), \textsc{\model} can improve CatE by an evident margin on term accuracy through integrating multiple types of signals. (2) On Yelp, \textsc{\model} underperforms CatE in terms of NPMI but significantly outperforms CatE in terms of P@20 and NDCG@20. Note that NPMI is an automatically computed metric, and the other two metrics rely on human annotation. Indeed, a recent study \cite{hoyle2021automated} shows that automatic metrics such as NPMI may not align well with human evaluation. From this perspective, we claim that \textsc{\model} performs better than CatE on Yelp, and our qualitative analysis below will validate this claim.

Besides quantitative evaluation, we show the qualitative comparison in Table \ref{tab:case}. We randomly select two seeds from NYT-Location, NYT-Topic, Yelp-Food, and Yelp-Sentiment, respectively. For each seed, we show top-5 terms retrieved by each method. A term is marked as incorrect ($\times$) if and only if at least 3 of the 5 annotators judge the term as irrelevant to the seed. Table \ref{tab:case} demonstrates that: (1) SeededLDA, Anchored CorEx, and KeyETM tend to find irrelevant or very general terms. For example, both Anchored CorEx and KeyETM retrieve the term ``\textit{fish}'' under the seed ``\textit{sushi}'', but ``\textit{fish}'' is also relevant to the seed ``\textit{seafood}'', thus it does not discriminatively belong to the sushi category. (2) Most terms discovered by CatE are accurate. However, CatE still makes mistakes in all four dimensions in Table \ref{tab:case}.
In contrast, \textsc{\model} achieves higher accuracy. If we further check the mistakes made by CatE, we can find general terms such as ``\textit{also}'' and ``\textit{savoury}'', which may co-occur frequently with other top-ranked terms. This possibly explains why CatE achieves higher NPMI than \textsc{\model} on Yelp since NPMI is based on the co-occurrence of retrieved terms.

\subsection{Ablation Study}
\begin{table}[t]
\small
\centering
\caption{Ablation study on different types of context information used in \textsc{\model}.}
\vspace{-0.5em}
\scalebox{0.925}{
	\begin{tabular}{p{3.2cm}<{\centering} | p{0.95cm}<{\centering} c | p{0.95cm}<{\centering} c}
		\toprule
		\multirow{2}{*}{Method} & \multicolumn{2}{c|}{\textbf{Yelp}-Food} & \multicolumn{2}{c}{\textbf{Yelp}-Sentiment} \\
		& P@20 & NDCG@20 & P@20 & NDCG@20 \\
		\midrule
		\textsc{\model} & \textbf{0.7912} & \textbf{0.8379} & \textbf{0.9750} & \textbf{0.9811}  \\
		\textsc{\model}-NoEmb & 0.4488 & 0.5335 & 0.9550 & 0.9646  \\
		\textsc{\model}-NoPLM & 0.6962 & 0.7602 & 0.7550 & 0.8029  \\
		\textsc{\model}-NoSntn & 0.7488 & 0.8029 & 0.9500 & 0.9631  \\
		\bottomrule
	\end{tabular}
}
\label{tab:ablation}
\vspace{-0.5em}
\end{table}

\begin{figure}[t]
\centering
\subfigure[\textbf{NYT}-Topic]{
\includegraphics[width=0.225\textwidth]{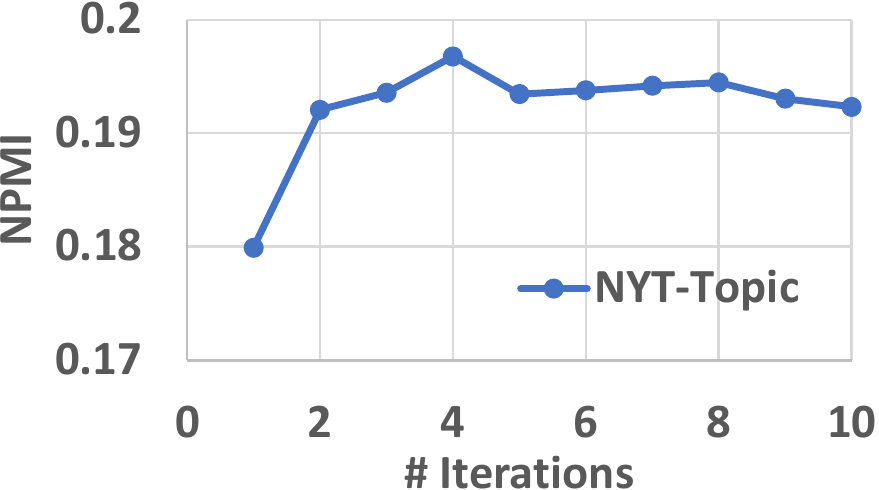}}
\hspace{0mm}
\subfigure[\textbf{NYT}-Location]{
\includegraphics[width=0.225\textwidth]{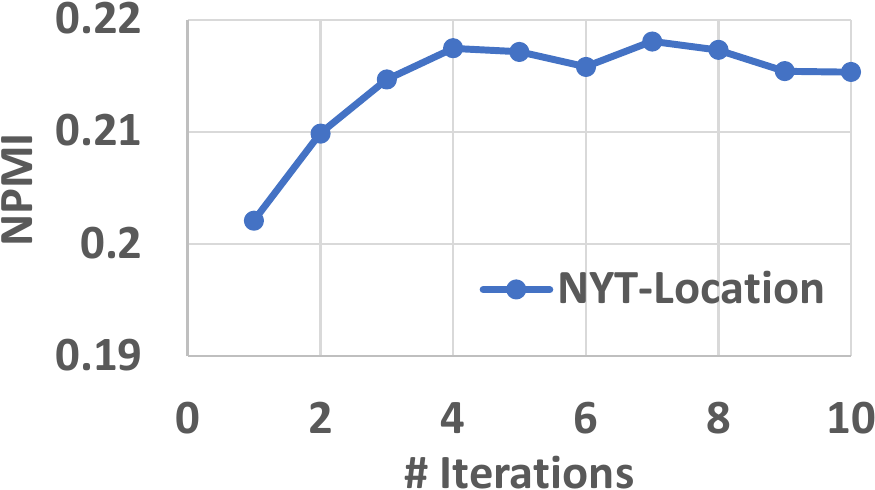}} \\
\subfigure[\textbf{Yelp}-Food]{
\includegraphics[width=0.225\textwidth]{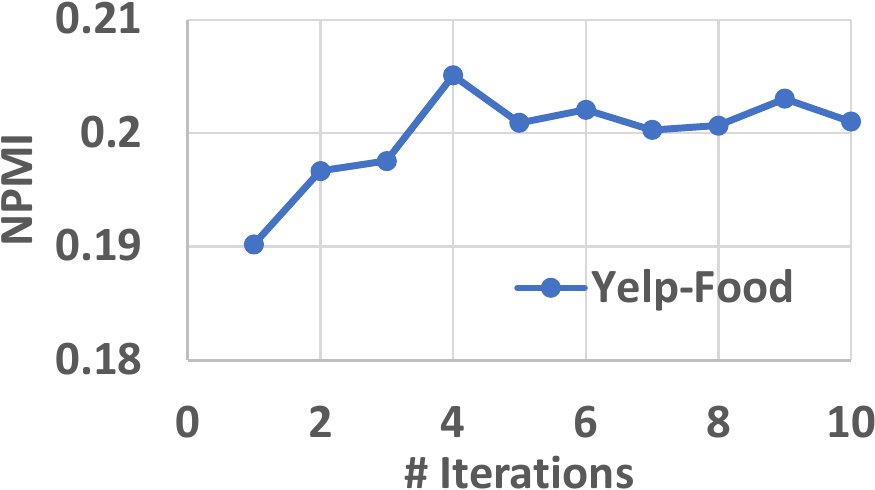}}
\hspace{0mm}
\subfigure[\textbf{Yelp}-Sentiment]{
\includegraphics[width=0.225\textwidth]{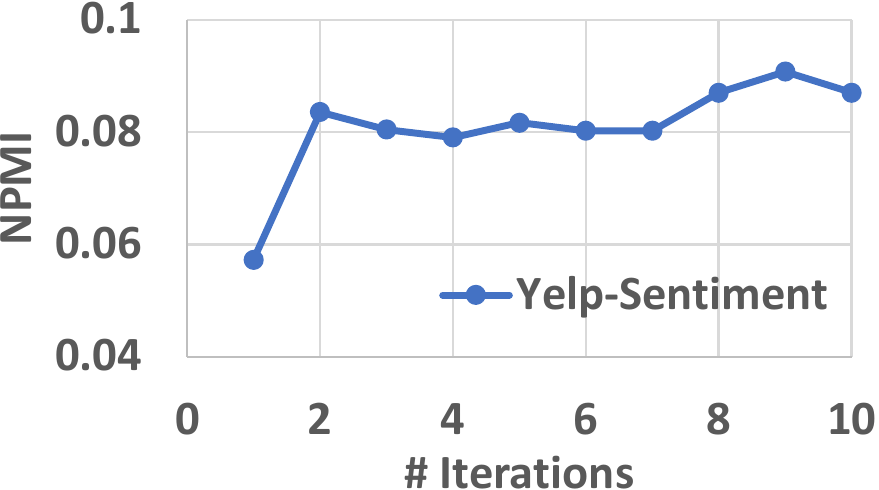}}
\vspace{-1em}
\caption{Effect of the number of iterations ($N$) on NPMI.}
\vspace{-1em}
\label{fig:iter}
\end{figure}

One key design in \textsc{\model} is the ensemble of multiple types of contexts. Specifically, we utilize the context information from three sources: seed-guided text embeddings (\textbf{Emb}), pre-trained language model representations (\textbf{PLM}), and topic-indicative sentences (\textbf{Sntn}). 
Now, we validate their contribution to the whole framework through an ablation analysis. Specifically, we can ignore one of the three sources while keeping all the other modules unchanged. This yields three ablation versions: \textbf{\textsc{\model}-NoEmb}, \textbf{\textsc{\model}-NoPLM}, and \textbf{\textsc{\model}-NoSntn}. 

Table \ref{tab:ablation} demonstrates the term accuracy scores of the full \textsc{\model} model and the three ablation versions. We can observe that: (1) \textsc{\model} consistently outperforms all three ablation versions, which implies the positive contribution of the three types of context signals. (2) Even for the same dataset (i.e., Yelp), the contribution of a certain type of context information varies significantly with the input seeds. For example, in the food dimension, \textsc{\model}-NoEmb performs the worst, which indicates that seed-guided embeddings are the most helpful signals. Meanwhile, in the sentiment dimension, the contribution of embeddings becomes the smallest. In comparison, pre-trained language model representations have the largest offering. This observation validates the motivation of this work that each type of context information has its specific value and limitation in topic discovery. None of them can dominate the others across all dimensions. Therefore, it becomes necessary to integrate them together, and our results show that the integration does achieve consistently the best performance.

\setlength{\tabcolsep}{3pt}
\begin{table*}[t]
\small
\centering
\caption{Extended qualitative results. $\times$: At least 3 of the 5 annotators judge the term as irrelevant to the seed.}
\vspace{-0.5em}
\scalebox{0.925}{
\begin{tabular}{c|c|l}
\toprule
\textbf{Dataset} & \textbf{Method} & \multicolumn{1}{c}{\textbf{Lower-ranked Terms}} \\
\midrule
\multirow{4}{*}{\textbf{NYT}-Topic} & 
CatE & 
\textbf{sports}: baseball, football, \wrong{clubs}, tennis, coaches, \wrong{amateur}, n.b.a, handball\\
& \textsc{\model} & 
\textbf{sports}: coaches, athletics, players, championships, sportsman, olympians, sporting events, tournament \\
& CatE & 
\textbf{politics}: \wrong{rhetoric}, \wrong{constituencies}, \wrong{vitriolic}, \wrong{passivity}, \wrong{unprincipled}, \wrong{polarized}, \wrong{philosophically}, \wrong{worldview} \\
& \textsc{\model} & 
\textbf{politics}: democratic, parties, conservative coalition, elected, liberal, electoral, \wrong{leaders}, political alliance \\
\midrule
\multirow{4}{*}{\textbf{Yelp}-Food} & 
CatE & 
\textbf{desserts}: churros, chocolate, \wrong{omelettes}, crepes, \wrong{truffles}, \wrong{fondue}, sweets, \wrong{breakfasts} \\
& \textsc{\model} & 
\textbf{desserts}: candied, scones, \wrong{truffles}, tarts, crepes, \wrong{coffees}, doughnuts, candies \\
& CatE & 
\textbf{seafood}: oysters, softshell, paella, fishes, octopus, mussel, mackerel, crawfish \\
& \textsc{\model} & 
\textbf{seafood}: lobster, clam, seafood, crawfish, blue crab, imitation crab, jumbo shrimp, sardines \\

\bottomrule
\end{tabular}
}
\vspace{-0.5em}
\label{tab:extended_case}
\end{table*}

\subsection{Parameter Study}
Another key design in \textsc{\model} is the iterative framework. To verify the contribution of multiple iterations, we conduct a parameter study by showing the NPMI of the discovered topics if we run \textsc{\model} for different numbers of iterations (i.e., $N$). Figure \ref{fig:iter} shows the effect of $N$ on NPMI across all four dimensions. 

From Figure \ref{fig:iter}, we see that: (1) When $N$ is small (e.g., $N \leq 4$), NPMI increases with $N$ in most cases. When we run \textsc{\model} for only one iteration, the performance is always significantly lower than that when we run 3-4 iterations. This finding validates our design choice that iteratively revising each topic can boost the topic coherence. (2) When $N$ becomes larger, the NPMI curve starts to fluctuate, and the performance gain of running more iterations is subtle. Moreover, more iterations will result in longer running time. Therefore, we believe that setting $N=4$ strikes a good balance.

\subsection{Case Study}
We have shown the top-5 terms retrieved by different algorithms in Table \ref{tab:case}. One may ask about the quality of lower-ranked terms in each topic. Thus, we conduct an extended case study by showing the 8 lowest-ranked terms among the top 20. These terms are listed in Table \ref{tab:extended_case}. Due to space limit, we only show the results of our \textsc{\model} model and the strongest baseline CatE, and two topics are selected for NYT-Topic and Yelp-Food, respectively.

From Table \ref{tab:extended_case}, we observe that the accuracy of CatE deteriorates for lower-ranked terms. For example, under the ``\textit{politics}'' seed from NYT-Topic, all of the 8 shown terms discovered by CatE are judged as irrelevant. By contrast, \textsc{\model} only makes one mistake under the same seed. This observation implies that the efficacy of \textsc{\model} can be generalized to the relatively lower part of the retrieved term list, which also reflects the robustness of \textsc{\model} by integrating multiple types of contexts.

\section{Related Work}

\noindent \textbf{Seed-Guided Topic Discovery.}
Different from supervised topic models (e.g., \cite{lacoste2008disclda,mcauliffe2007supervised}) that rely on a large number of human-annotated documents, seed-guided topic discovery only requires a set of user-interested seeds to find corresponding topics. In \cite{andrzejewski2009incorporating}, seeds are incorporated as prior of topic modeling using must-link and cannot-link constraints. SeededLDA \cite{jagarlamudi2012incorporating} uses seeds to bias topics to produce seed terms and documents to select topics containing them. Anchored CorEx \cite{gallagher2017anchored} discovers informative topics with correlation maximization and leverages seeds by balancing corpus compression and seed-indicative information. Recent studies also incorporate embedding learning techniques to obtain more accurate semantic representations. For instance, CatE \cite{meng2020discriminative} learns category-guided text embeddings by enforcing distinctiveness among seeds in the embedding space; SeeTopic~\cite{zhang2022seed} further utilizes the power of pre-trained language models for better text representations and the ability to handle out-of-vocabulary seeds.

\vspace{1mm}

\noindent \textbf{Representation-Enhanced Topic Discovery.} 
With the rapid development in text representation learning, recent topic discovery methods incorporate distributed representations to enhance the modeling of text semantics.
Earlier approaches incorporate context-free word embeddings \cite{mikolov2013distributed} into classic probabilistic topic models (e.g., LDA \cite{blei2003latent}), including Gaussian LDA \cite{das2015gaussian}, LFTM \cite{nguyen2015improving}, Spherical HDP \cite{batmanghelich2016nonparametric}, and CGTM \cite{xun2017correlated}. 
TWE \cite{liu2015topical} learns embeddings based on associations between words and latent topics obtained by LDA. CLM \cite{xun2017collaboratively} collaboratively models topics and learns word embeddings by considering both global and local contexts. ETM \cite{dieng2020topic} learns topic embeddings in the word embedding space to improve LDA for a better fit of a large vocabulary. 
More recent studies leverage the contextualized representations generated by pre-trained language models (e.g., BERT \cite{devlin2019bert}) to facilitate the discovery of coherent topics.
These contextualized representations can be used either at token-level for clustering to form topics \cite{meng2022topic,sia2020tired,thompson2020topic,zhang2022neural} or at document-level for modeling document-topic correlations \cite{bianchi2021pre,grootendorst2022bertopic}.

\section{Conclusions and Future Work}
In this work, we study seed-guided topic discovery by learning from multiple types of contexts, including skip-gram embeddings based on local contexts, pre-trained language model representations upon general-domain pre-training, and topic-indicative sentences retrieved according to seed-distinctive terms.
Our proposed \textsc{\model} framework jointly leverages these contexts via an ensemble process for robust topic discovery under different types of seeds.
On two real-world datasets and across four sets of seeds, \textsc{\model} consistently outperforms existing seed-guided topic discovery approaches in terms of topic coherence and term accuracy.

For future studies, the promising topic discovery results achieved by \textsc{\model} may further benefit keyword-based text classification \cite{mekala2020contextualized,zhang2019higitclass} via expanding the seed word semantics and prompt-based methods \cite{schick2021exploiting} via enriching their verbalizers. Also, \textsc{\model} can be extended to model input seeds organized in a hierarchical manner by injecting hierarchy regularization or discovering topics beyond the provided seeds by incorporating latent topic learning in the corpus modeling process.

\section*{Acknowledgments}
We thank Yichen Liu and Ruining Zhao for their help with annotation and anonymous reviewers for their valuable and insightful feedback.
Research was supported in part by the IBM-Illinois Discovery Accelerator Institute, US DARPA KAIROS Program No. FA8750-19-2-1004 and INCAS Program No. HR001121C0165, National Science Foundation IIS-19-56151, IIS-17-41317, and IIS 17-04532, and the Molecule Maker Lab Institute: An AI Research Institutes program supported by NSF under Award No. 2019897, and the Institute for Geospatial Understanding through an Integrative Discovery Environment (I-GUIDE) by NSF under Award No. 2118329. Any opinions, findings, and conclusions or recommendations expressed herein are those of the authors and do not necessarily represent the views, either expressed or implied, of DARPA or the U.S. Government.

\balance
\bibliography{wsdm}

\end{spacing}
\end{document}